\documentclass[journal,twocolumn,10pt,letter]{IEEEtran}
\IEEEoverridecommandlockouts
\usepackage{cite}
\usepackage{amsmath,amssymb,amsfonts}
\usepackage{algorithmicx}
\usepackage{graphicx}
\usepackage{textcomp}
\usepackage{xcolor}
\usepackage{booktabs}
\usepackage{multirow}
\usepackage{array}
\usepackage{tabularx}
\usepackage{cite}
\usepackage{url}
\usepackage{hyperref}
\usepackage{algorithm}
\usepackage{algpseudocode}

\def\BibTeX{{\rm B\kern-.05em{\sc i\kern-.025em b}\kern-.08em T\kern-.1667em\lower.7ex\hbox{E}\kern-.125emX}}

\begin{document}

\title{GraphTrafficGPT: Enhancing Traffic Management through Graph-Based AI Agent Coordination}


\author{\IEEEauthorblockN{Nabil Abdelaziz Ferhat Taleb\IEEEauthorrefmark{1},
Abdolazim Rezaei\IEEEauthorrefmark{1},
Raj Atulkumar Patel\IEEEauthorrefmark{1},  and
Mehdi Sookhak\IEEEauthorrefmark{1} 
}\\
\IEEEauthorblockA{\IEEEauthorrefmark{1}Department of Computer Science,
Texas A\&M University, Corpus Christi, TX 78412 USA}
}


\maketitle

\begin{abstract}
Large Language Models (LLMs) offer significant promise for intelligent traffic management; however, current chain-based systems like TrafficGPT are hindered by sequential task execution, high token usage, and poor scalability, making them inefficient for complex, real-world scenarios. To address these limitations, we propose GraphTrafficGPT, a novel graph-based architecture, which fundamentally redesigns the task coordination process for LLM-driven traffic applications. GraphTrafficGPT represents tasks and their dependencies as nodes and edges in a directed graph, enabling efficient parallel execution and dynamic resource allocation. The main idea behind the proposed model is a Brain Agent that decomposes user queries, constructs optimized dependency graphs, and coordinates a network of specialized agents for data retrieval, analysis, visualization, and simulation. By introducing advanced context-aware token management and supporting concurrent multi-query processing, the proposed architecture handles interdependent tasks typical of modern urban mobility environments.  Experimental results demonstrate that GraphTrafficGPT reduces token consumption by 50.2\% and average response latency by 19.0\% compared to TrafficGPT, while supporting simultaneous multi-query execution with up to 23.0\% improvement in efficiency.

\end{abstract}


 \section{Introduction}
Large Language Models (LLMs) have changed artificial intelligence capabilities across domains by enabling natural language understanding and generation at new levels. The recent models, such as GPT-4, Claude, and Llama, can comprehend complex instructions, reason through problems, and generate coherent responses across diverse applications~\cite{wu2023}. Their emergence has changed how humans interact with computational systems, replacing command interfaces with natural language conversations.

In the transportation domain, the complexity of urban mobility systems has created challenges for traffic management and optimization. Modern transportation networks generate vast amounts of heterogeneous data from sensors, cameras, and connected vehicles that must be analyzed and acted upon efficiently~\cite{zheng2023horizon}. Traditional traffic management approaches often struggle with data integration, intuitive interfaces, and real-time analysis capabilities necessary for responsive urban mobility systems.

The integration of LLMs and transportation systems offers opportunities to address these challenges. Recent researches have explored applications of LLMs in traffic management, including accident analysis~\cite{wu2024}, traffic signal optimization~\cite{lai2024}, and simulation scenario generation~\cite{tan2023}. These approaches use the natural language capabilities of LLMs to create accessible interfaces for traffic management professionals.

LLMs have demonstrated remarkable potential in transforming transportation systems through multiple avenues of application. Zheng et al.~\cite{zheng2023safety} proposed a framework that leverages LLMs to enhance traffic safety by generating automated crash reports and analyzing multi-modal sensor data from roadside units (RSUs) and vehicles to provide proactive safety recommendations. Li et al.~\cite{li2024} utilized LLMs to interpret complex transportation scenarios and convert them into executable simulations in SUMO, enabling traffic planners to explore design alternatives through conversational interfaces. In~\cite{da2023}, an LLM-based approach was introduced for traffic signal control that bridges simulation and real-world deployment to facilitate more efficient coordination during congestion events.

The integration of LLMs with simulation environments has also shown promising results. Villarreal et al.~\cite{villarreal2023} demonstrated how ChatGPT can increase the success rate of policies in mixed traffic control tasks, making these tools more accessible to operators without specialized expertise. Similarly, in \cite{wang2024}, an LLM-driven traffic performance system was developed that combines historical data with real-time observations to generate insights for transportation surveillance and management. These applications demonstrate how LLMs can make transportation systems more accessible, efficient, and responsive to changing conditions.

Despite these advancements, most current LLM applications in transportation use chain-based processing architectures, where information flows sequentially through predetermined steps. While effective for simple tasks, this approach presents significant limitations for complex transportation scenarios that require parallel processing and dynamic adaptation. Among the various frameworks, TrafficGPT~\cite{zhang2023} represents one of the most comprehensive attempts to integrate LLMs with specialized Traffic Foundation Models (TFMs) for transportation management and traffic data analysis, making it an ideal candidate for further enhancement and optimization. By combining the natural language understanding of LLMs with the capabilities of traffic modeling tools, TrafficGPT enables users to analyze traffic data, visualize patterns, and support decision-making through natural language interaction.

However, the original TrafficGPT implementation has several limitations that impede its practical deployment: \textbf{firstly}, the chain-based reasoning approach requires repeated context loading and sequential processing, leading to excessive token usage and higher operational costs; \textbf{secondly}, tasks that could be executed in parallel are processed sequentially, resulting in longer response times and reduced system throughput; \textbf{thirdly}, the linear nature of chain-based processing restricts efficient information sharing between different components of the system; and \textbf{finally}, the original architecture can process only one query at a time, limiting its practical utility in real-world traffic management scenarios.

To address these limitations, we propose \textbf{GraphTrafficGPT}, a graph-based architecture that transforms the sequential chain into a dependency graph with parallel execution capabilities. Our approach fundamentally reimagines how language models interact with traffic management systems by replacing linear processing with a multi-dimensional graph structure that more closely resembles how human experts approach complex transportation problems. The system identifies dependencies between subtasks and generates an optimized execution plan that maximizes parallel processing while maintaining necessary sequential relationships.

GraphTrafficGPT employs a hybrid framework that combines the natural language understanding capabilities of LLMs with the specialized knowledge of TFMs through a graph-based coordination mechanism. By representing traffic management tasks as nodes in a dependency graph, the system can identify independent operations that can be executed concurrently, significantly reducing response times for complex queries. Additionally, the architecture implements context-aware token management techniques that eliminate redundant information processing, reducing operational costs while maintaining or improving response quality. Through dynamic resource allocation and parallel execution paths, GraphTrafficGPT achieves substantially higher throughput than chain-based alternatives, especially for multi-part queries that characterize real-world traffic management scenarios.

The contributions of this paper are as follows:

\begin{itemize}
    \item We introduce a novel approach that represents each traffic management task as a node in a directed graph, where edges indicate dependencies. This graph-based structure enables the system to identify independent tasks for parallel execution, while ensuring correct sequencing for dependent operations—addressing fundamental scalability and efficiency limitations of prior chain-based models.
    \item We design a centralized Brain Agent that analyzes user queries, decomposes them into discrete tasks, identifies inter-task dependencies, and orchestrates workflow execution across a network of specialized agents. The Brain Agent also maintains a global context, effectively reducing redundant information processing across tasks.
    \item  We develop a modular network of domain-specific agents—including Data Retrieval, Traffic Analysis, Visualization, Simulation, and Workflow Agents—each equipped with a ReAct (Reasoning and Action) loop for iterative problem-solving and seamless integration with Traffic Foundation Models.
    \item We propose and implement advanced context pruning and sharing techniques to minimize token usage. By eliminating redundant information and selectively propagating relevant context between related tasks, our system significantly improves computational efficiency and cost-effectiveness.
    \item By identifying shared dependencies and coordinating execution across the agent network, our architecture supports the simultaneous processing of multiple related queries—overcoming the strictly sequential handling of traditional frameworks.
\end{itemize}

The experimental results show improvements over the original TrafficGPT implementation, including a 50\% reduction in token consumption, a 19\% decrease in latency, and support for concurrent multi-query processing with up to 37.6\% improved efficiency. These changes make GraphTrafficGPT practical for real-world deployment in traffic management centers and integrated urban systems.

The rest of the paper is organized as follows. Section II presents a comprehensive review of related work on Large Language Models in transportation and the evolution from chain-based to graph-based architectures. Section III details our methodology, including the system architecture, graph-based task decomposition, and multi-query support mechanisms. Section IV describes our experimental setup and presents comparative results with the original TrafficGPT implementation. Section V discusses the implications of our findings and potential applications in real-world traffic management scenarios. Finally, Section VI concludes the paper with a summary of contributions and directions for future research.

\section{Related Work}

\subsection{Large Language Models in Transportation}
Recent research has witnessed significant growth in applying Large Language Models (LLMs) to transportation systems. A comprehensive survey by Nie et al. highlights how LLMs are transforming transportation through capabilities like language understanding, in-context learning, and multimodal reasoning~\cite{nie2024}. These advancements address inherent limitations in traditional transportation systems by processing unstructured data and encoding domain-specific knowledge.

Several recent studies have explored LLM applications in traffic safety. Zheng et al. investigated ChatGPT's potential for accident report generation and traffic data analysis, demonstrating how LLMs with cross-modal encoders can process crash-related information from text, images, and audio inputs~\cite{zheng2023horizon}. Building on this work, Zheng et al.~\cite{zheng2023safety} developed TrafficSafetyGPT, a specialized model fine-tuned on transportation safety guidelines to provide domain-specific expertise. 

For traffic signal optimization, Da et al.~\cite{da2023}introduced PromptGAT, integrating LLM-based dynamics modeling with reinforcement learning to enhance traffic signal control. Their framework leverages domain knowledge and real-time traffic data to predict system dynamics and improve policy adaptation. Similarly, the work presented in~\cite{villarreal2023}  demonstrates that ChatGPT significantly increases the success rate of policies in mixed traffic control tasks. 

In the realm of traffic data management, Wang et al.~\cite{wang2024} developed Traffic Performance GPT (TP-GPT), which uses LLMs to generate SQL queries and interpret traffic data from real-time databases. This approach enables transportation analysts to interact with complex databases through natural language, significantly improving accessibility. For traffic simulation, Li et al.~\cite{li2024} introduced ChatSUMO, which integrates LLM capabilities with the SUMO traffic simulator. This system transforms textual descriptions into executable simulations, making traffic modeling accessible to users without specialized knowledge. 

Perhaps most relevant to our work, Zhang et al.~\cite{zhang2023} proposed TrafficGPT, a framework that integrates LLMs with specialized Traffic Foundation Models (TFMs). While this approach enables users to analyze traffic data and support decision-making through natural language interaction, its chain-based architecture introduces inefficiencies in token usage and processing time that limit practical deployment at scale.

\subsection{Graph-Based Approaches vs. Chain-Based Methods}
The architectural design of LLM-powered systems significantly impacts their efficiency, scalability, and performance. Current approaches can be broadly categorized into chain-based and graph-based methods, each with distinct characteristics and capabilities. Chain-based approaches, exemplified by the Chain-of-Thought (CoT) prompting introduced by Wei et al. \cite{wei2022}, process information sequentially, with each component receiving input solely from its predecessor. While this approach has improved LLM reasoning capabilities, the authors in~\cite{yao2022} identified limitations in its linear processing when handling complex reasoning involving iterative thinking or multiple knowledge sources. 

To address these limitations, Yao et al.~\cite{yao2022} proposed ReAct, which interleaves reasoning traces with actions in an iterative manner. Their experiments demonstrated improved performance on knowledge-intensive tasks by allowing models to retrieve and incorporate external information during the reasoning process. However, even this approach maintains a fundamentally sequential processing paradigm. 

Graph-based architectures have demonstrated superior performance across various LLM applications. Weng et al.~\cite{weng2023} proposed Graph of Thoughts (GoT), a framework that models each thought generated by an LLM as a node within a graph, with edges representing dependencies between thoughts. This approach enables sophisticated control structures including loops, branches, and parallel execution paths. As noted in their paper, ``When working on a novel idea, a human would not only follow a chain of thoughts (as in CoT) or try different separate ones (as in ToT), but would actually form a more complex network of thoughts.''

Microsoft Researchers in~\cite{microsoft2024} introduced GraphRAG, a system that uses LLM-generated knowledge graphs to improve question-answering performance when analyzing complex information. Their technical report demonstrated significant improvements in correctness and contextual alignment compared to traditional retrieval methods. Xia et al.~\cite{xia2024} conducted a comparative study of graph-based versus sequential approaches for traffic simulation with LLMs, finding that graph-based architectures scale more effectively with increasing task complexity, with performance gaps widening as the number of interdependent subtasks grows. This is particularly relevant for traffic management systems that must coordinate multiple interrelated functions. 

As highlighted in \cite{dong2024}, graph-based approaches offer several advantages over chain-based methods, including: (1) Parallel Execution: Tasks can be executed concurrently based on the dependency graph, enabling efficient resource utilization; (2) Lower Token Usage: Redundant steps are avoided by resolving dependencies upfront and sharing context efficiently between nodes; (3) Faster Response Time: Batch/parallel execution reduces overall latency, particularly for complex multi-step tasks; (4) Lower Error Rate: Dependencies are resolved before execution, reducing cascading errors that commonly occur in linear processing; and (5) Better Scalability: Complex interdependencies are handled more efficiently through explicit modeling in the graph structure.

While existing works have made significant progress in applying LLMs to transportation tasks and developing graph-based architectures for LLM systems separately, there remains a critical gap in combining these approaches for practical traffic management applications. Current LLM-based traffic management systems predominantly rely on chain-based architectures that suffer from high token consumption, sequential execution bottlenecks, limited context handling, and single-query restrictions. These limitations severely constrain their practical utility in real-world traffic management centers where operators must monitor and respond to multiple aspects of transportation systems simultaneously. Furthermore, the increasing complexity of urban mobility networks demands more efficient and scalable solutions capable of processing interdependent traffic management tasks in parallel. To address these issues, this paper proposes GraphTrafficGPT, a graph-based architecture that transforms the sequential chain-based approach of TrafficGPT into a dependency graph with parallel execution capabilities, enabling more efficient and responsive traffic management.

\section{Methodology}
\subsection{System Overview}

GraphTrafficGPT maintains the core functionality of the original TrafficGPT system while fundamentally redesigning its architecture. The system integrates LLMs with specialized Traffic Foundation Models (TFMs) through a graph-based coordination mechanism that enables more efficient processing and parallel execution.

The overall architecture of GraphTrafficGPT consists of four main components:

\begin{enumerate}
\item \textbf{Input Processing Module}: Analyzes user queries and decomposes them into discrete tasks.
\item \textbf{Dependency Graph Generator}: Identifies relationships between tasks and constructs an optimized execution graph.
\item \textbf{Brain Agent}: Coordinates task distribution and execution across specialized agents.
\item \textbf{Response Integration Module}: Combines outputs from multiple TFMs into coherent responses.
\end{enumerate}

\begin{figure}[!t]
\centering
\includegraphics[width=\columnwidth]{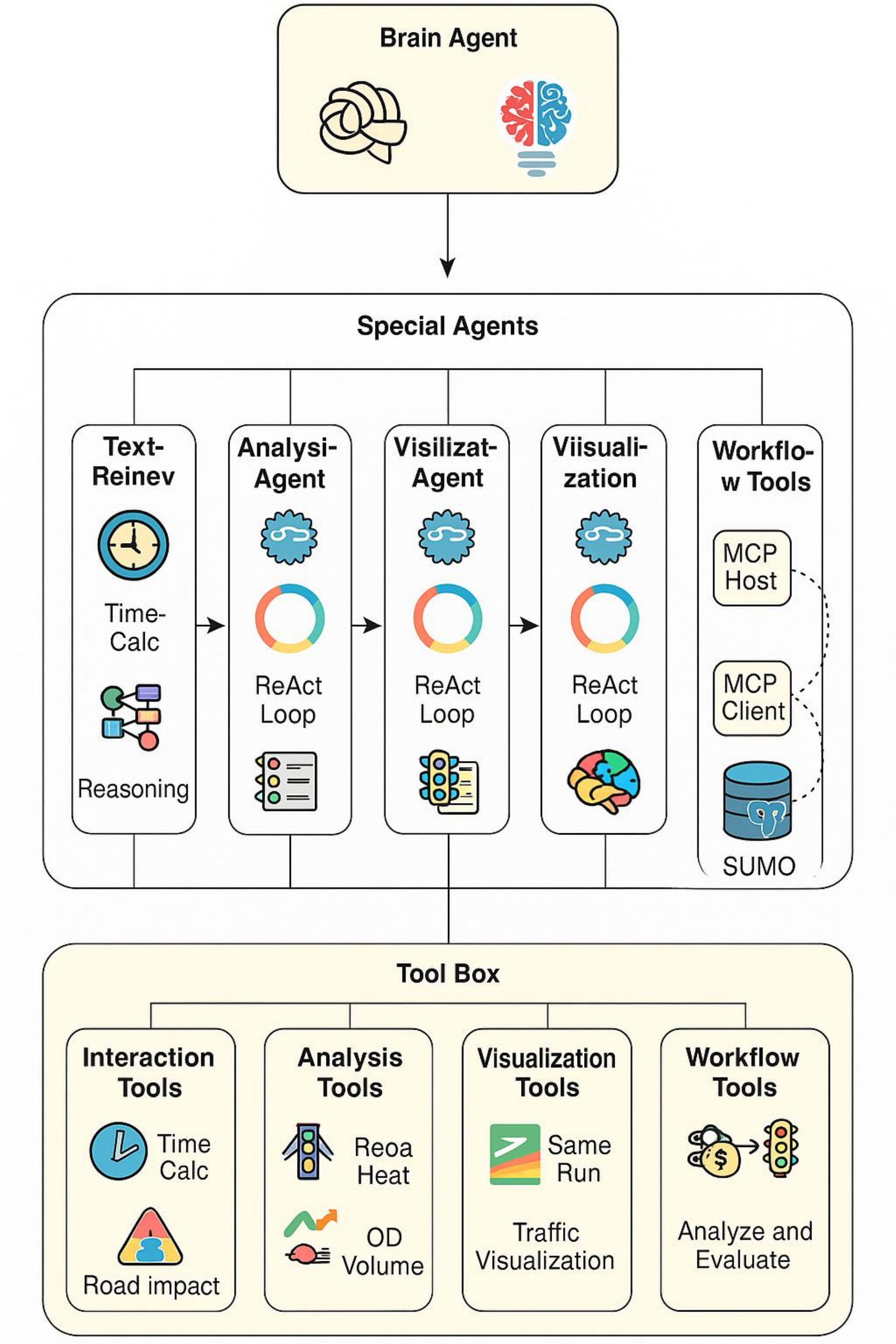}
\caption{GraphTrafficGPT Architecture Overview: The system processes user queries through a comprehensive pipeline that transforms natural language requests into parallel executable tasks using the graph-based approach. The architecture enables efficient dependency management, resource allocation, and multi-task processing, leading to significant improvements in token usage and response time.}
\label{fig_architecture}
\end{figure}

As illustrated in Figure \ref{fig_architecture}, the system's architecture enables the decomposition of complex traffic management tasks into optimized workflows that can be executed in parallel. This design represents a significant advancement over the sequential processing approach used in the original TrafficGPT implementation.

\subsection{Brain Agent and Specialized Agents}

\begin{figure}[!t]
\centering
\includegraphics[width=\columnwidth]{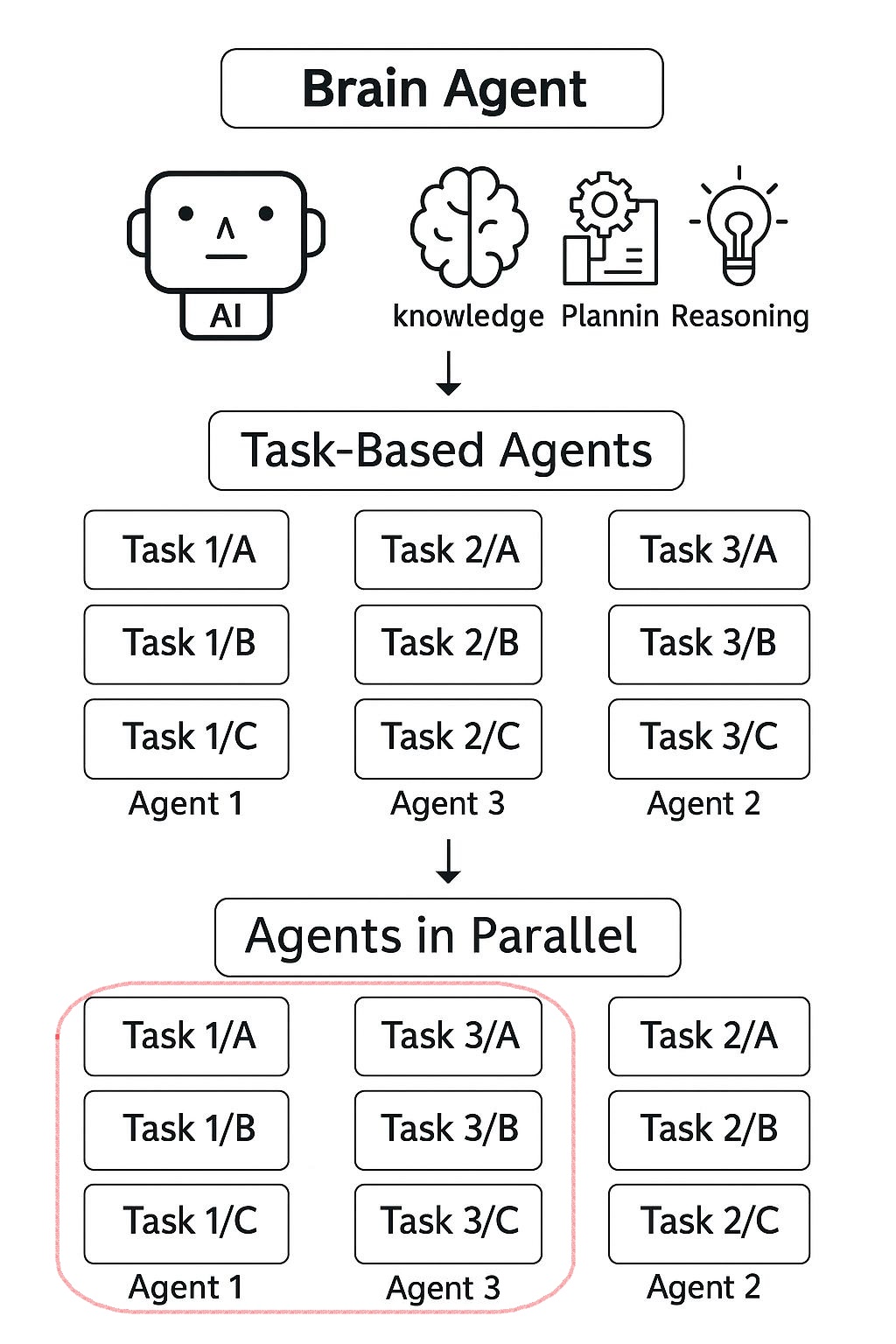}
\caption{Brain Agent and Multi-Task Parallelization: The figure illustrates how GraphTrafficGPT's brain agent coordinates multiple tasks across different specialized modules. Each specialized agent contains a ReAct loop for handling complex tasks, a knowledge component, and reasoning capabilities. The system maintains context across tasks while distributing processing across parallel execution paths, enabling handling of complex, multi-part queries with significantly reduced latency compared to sequential approaches.}
\label{fig_brain_agent}
\end{figure}

At the heart of GraphTrafficGPT lies the Brain Agent, which serves as the central coordinator of the system. As shown in Figure \ref{fig_brain_agent}, the Brain Agent performs several critical functions:

\begin{enumerate}
\item \textbf{Task Decomposition}: Analyzes user inputs and breaks them down into discrete tasks.
\item \textbf{Dependency Analysis}: Identifies which tasks depend on others and which can be executed independently.
\item \textbf{Agent Assignment}: Assigns tasks to specialized agents based on task type and requirements.
\item \textbf{Execution Coordination}: Monitors execution and manages the flow of information between agents.
\end{enumerate}

GraphTrafficGPT employs multiple specialized agents, each dedicated to a specific domain of traffic management:

\begin{enumerate}
\item \textbf{Data Retrieval Agent}: Handles access to traffic databases and time-related queries.
\item \textbf{Traffic Analysis Agent}: Processes traffic volume and performance data.
\item \textbf{Visualization Agent}: Creates visual representations of traffic information.
\item \textbf{Simulation Agent}: Manages traffic simulation tasks.
\item \textbf{Workflow Agent}: Handles optimization and trend analysis.
\item \textbf{General Query Agent}: Processes non-specialized traffic-related queries.
\end{enumerate}

The coordination between these specialized agents is facilitated by a Multi-Agent Communication Protocol (MCP) that ensures efficient information exchange. Each agent operates as an MCP Client, connecting to a centralized MCP Host that manages message routing and synchronization. This protocol enables asynchronous communication, context sharing, and real-time coordination of task dependencies, which is essential for the parallel execution capabilities of GraphTrafficGPT.

Each agent incorporates a ReAct (Reasoning and Action) loop that enables iterative problem-solving for complex tasks. This loop allows agents to break down challenging problems into manageable steps, reason about intermediate results, and adjust their approach as needed. Knowledge components within each agent store domain-specific information, enhancing their capabilities for specialized tasks.

\subsection{Graph-Based Task Decomposition}

Unlike the chain-based approach used in TrafficGPT, GraphTrafficGPT represents tasks as nodes in a directed graph, with edges indicating dependencies between tasks. This representation enables the system to identify tasks that can be executed in parallel, as well as those that require sequential processing.

\begin{figure}[!t]
\centering
\includegraphics[width=0.8\columnwidth]{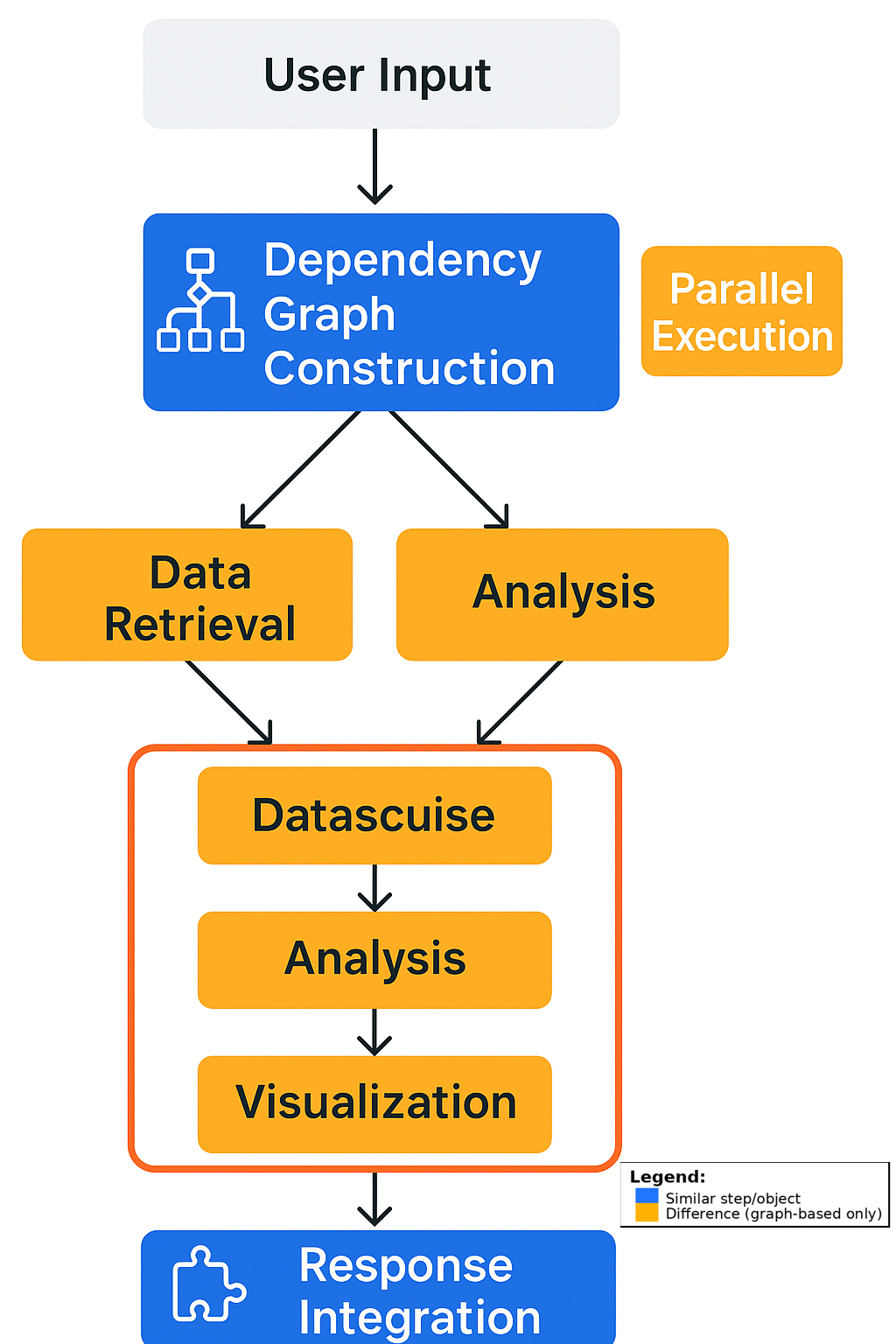}
\caption{Comparison of Graph-Based and Chain-Based Approaches: The graph-based approach (right) enables parallel execution and eliminates redundant operations present in the chain-based approach (left).}
\label{fig_graph_vs_chain}
\end{figure}

\begin{figure*}[!t]
\centering
\includegraphics[width=\textwidth]{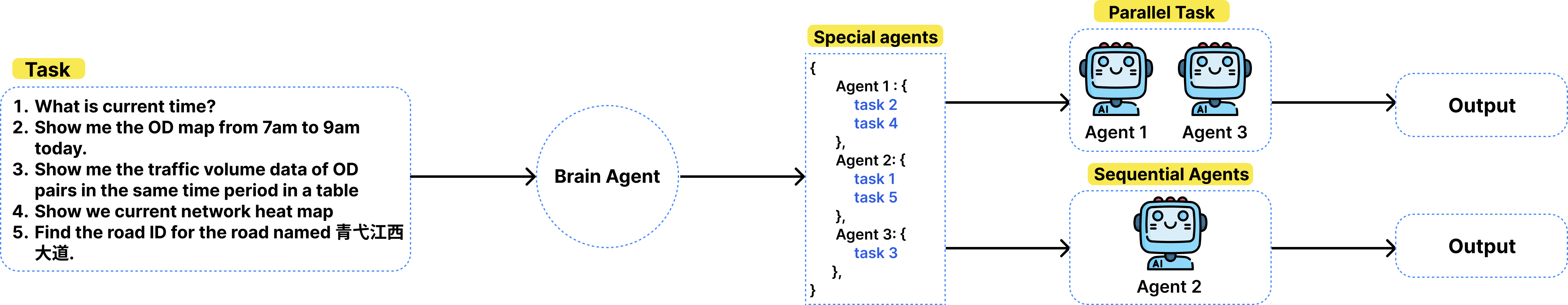}
\caption{Task Execution Workflow: This diagram illustrates how GraphTrafficGPT processes user inputs through its graph-based architecture. Multiple queries are parsed by the Brain Agent and distributed to appropriate specialized agents based on their dependencies. This enables parallel execution of independent tasks while ensuring proper sequencing for dependent operations, significantly reducing overall processing time.}
\label{fig_workflow}
\end{figure*}

The task decomposition process follows these steps:

\begin{enumerate}
\item The Brain Agent analyzes the user query to identify required tasks.
\item Each task is represented as a node in the dependency graph.
\item Tasks are grouped based on their dependencies and type.
\item Independent tasks are assigned to appropriate specialized agents for parallel execution.
\item Dependent tasks are scheduled to execute after their prerequisites are completed.
\item The resulting execution plan is optimized to maximize parallel processing.
\end{enumerate}

As shown in Figure \ref{fig_workflow}, this graph-based approach enables efficient task processing by identifying independent components that can be executed simultaneously. When tasks depend on the outputs of others, they are sequenced appropriately while still allowing unrelated tasks to proceed in parallel. This contrasts with the sequential approach illustrated in Figure \ref{fig_graph_vs_chain}, where each task must wait for its predecessor to complete before beginning execution, regardless of actual dependencies.

\subsection{System Algorithm}

\begin{algorithm}[!t]
\caption{GraphTrafficGPT System}
\label{alg:graphtrafficgpt}
\begin{algorithmic}[1] 

\Function{ProcessQuery}{$user\_query$}
    \State $tasks \gets$ BreakdownQuery($user\_query$)
    \State $graph \gets$ BuildDependencyGraph($tasks$)
    \State $results \gets$ [ ]
    
    \While{UnprocessedTasks($graph$)}
        \State $ready\_tasks \gets$ GetIndependentTasks($graph$)
        \ForAll{$task$ \textbf{in} $ready\_tasks$}  \Comment{Parallel execution}
            \State \quad $agent \gets$ SelectAgent($task.type$)
            \State \quad $context \gets$ GetPreviousContext($results$)
            \State \quad $result \gets$ agent.Execute($task, context$)
            \State \quad $results$.append($result$)
            \State \quad MarkComplete($graph, task$)
        \EndFor
    \EndWhile
    
    \State \Return CombineResults($results$)
\EndFunction

\vspace{0.5em}

\Function{SelectAgent}{$task\_type$}
    \If{$task\_type =$ DATA}
        \State \Return DataAgent
    \ElsIf{$task\_type =$ ANALYSIS}
        \State \Return AnalysisAgent
    \ElsIf{$task\_type =$ VISUAL}
        \State \Return VisualizationAgent
    \ElsIf{$task\_type =$ SIMULATION}
        \State \Return SimulationAgent
    \ElsIf{$task\_type =$ OPTIMIZE}
        \State \Return OptimizationAgent
    \Else
        \State \Return GeneralAgent
    \EndIf
\EndFunction

\end{algorithmic}
\end{algorithm}

To formalize the operation of GraphTrafficGPT, Algorithm~\ref{alg:graphtrafficgpt} presents the simplified pseudocode for the system workflow. The algorithm illustrates the key innovation of GraphTrafficGPT: parallel execution of independent tasks through dependency graph analysis. The system breaks down complex queries, identifies which tasks can run simultaneously, and distributes them across specialized agents while maintaining proper execution order for dependent tasks.

\subsection{Multi-Query Support and Parallel Execution}

A key innovation in GraphTrafficGPT is its ability to process multiple queries simultaneously. Unlike traditional systems that handle each query sequentially, GraphTrafficGPT can identify relationships between queries and process them in a coordinated manner. This is achieved through:

\begin{enumerate}
\item \textbf{Query Analysis}: Related queries are identified based on their dependencies and requirements.
\item \textbf{Parallel Agent Assignment}: Multiple specialized agents can work simultaneously on different aspects of the query set.
\item \textbf{Resource Allocation}: Computational resources are dynamically allocated based on task complexity and priority.
\item \textbf{Concurrent ReAct Loops}: Multiple reasoning and action loops operate in parallel across different agents.
\item \textbf{Shared Context Management}: Information relevant to multiple queries is stored in a centralized context accessible to all agents.
\end{enumerate}

Figure \ref{fig_brain_agent} illustrates how the Brain Agent orchestrates this multi-task parallelization. By distributing tasks across specialized agents operating in parallel, the system can process complex, multi-part queries far more efficiently than sequential architectures. Each agent can handle multiple subtasks through its ReAct loop, further enhancing processing efficiency.

When a task requires additional processing capabilities, the specialized agent calls upon specific tools from the Tool Box shown at the bottom of Figure \ref{fig_brain_agent}. These tools include various Traffic Foundation Models for data retrieval, analysis, visualization, simulation, and workflow optimization. The agent manages the interaction with these tools, processes their outputs, and integrates the results into its response.

\section{Experimental Results}
\subsection{Experimental Setup}

To evaluate the performance of GraphTrafficGPT compared to the original TrafficGPT implementation, we conducted experiments using a standardized set of traffic management queries across both systems. The experiments were performed using the same underlying TFMs and computational resources to ensure fair comparison.

The evaluation metrics included:
\begin{enumerate}
\item Token consumption (efficiency)
\item Response latency (speed)
\item Multi-query processing capability
\item Output quality and correctness
\end{enumerate}

\subsection{Token Usage Comparison}

\begin{figure}[!t]
\centering
\includegraphics[width=\columnwidth]{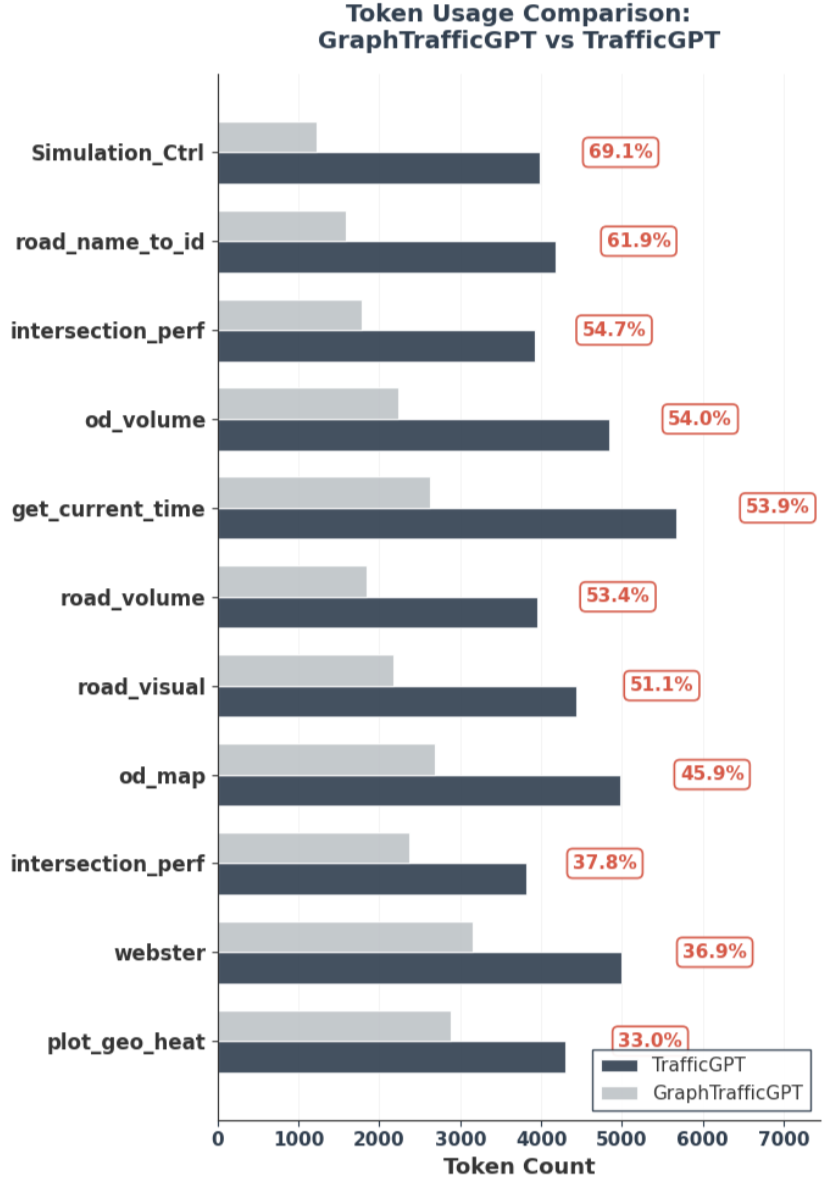}
\caption{Token Usage Comparison between GraphTrafficGPT and TrafficGPT: The light bars represent GraphTrafficGPT's token consumption while dark bars show TrafficGPT's usage. Percentage reductions are displayed for each function, demonstrating the efficiency gains achieved through our graph-based architecture.}
\label{fig:token_usage}
\end{figure}

Figure~\ref{fig:token_usage} presents a comprehensive comparison of token consumption between the original TrafficGPT and our improved GraphTrafficGPT implementation across various traffic management tasks. The horizontal bar chart visualizes the dramatic reduction in token usage achieved through our graph-based approach.

The results demonstrate an average token reduction of 50.2\% across all tasks, with some functions showing remarkable reductions exceeding 60\%. Notably, the \texttt{Simulation\_Controller} and \texttt{road\_name\_to\_id} functions achieved reductions of 69.1\% and 61.9\% respectively, highlighting the effectiveness of our approach in eliminating redundant processing for complex tasks. The graph-based architecture particularly excels in scenarios involving multiple interdependent operations, where traditional sequential processing would require repeated context loading.

\subsection{Latency Analysis}

\begin{figure}[!t]
\centering
\includegraphics[width=\columnwidth]{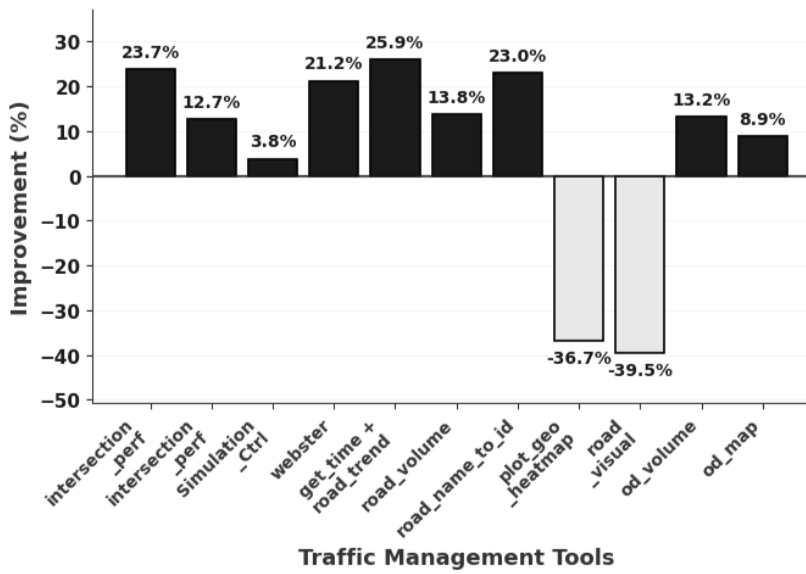}
\caption{Latency Analysis: Performance improvements and degradations for GraphTrafficGPT compared to TrafficGPT. The majority of functions show significant latency reductions, with \texttt{intersection\_performance} achieving a 23.7\% improvement.}
\label{fig:latency}
\end{figure}

The latency performance comparison, illustrated in Figure~\ref{fig:latency}, reveals the speed improvements achieved by GraphTrafficGPT across different traffic management functions. While most functions show positive improvements, the visualization also transparently presents cases where the graph-based approach introduces overhead.

The latency measurements show an average improvement of 19.0\% with GraphTrafficGPT. The most significant latency reduction was observed in intersection performance queries (23.7\%), where the graph-based approach enabled more efficient handling of contextually similar queries through intelligent context sharing and parallel execution. However, two visualization functions (\texttt{plot\_geo\_heatmap} and \texttt{road\_visualization}) experienced increased latency of 36.7\% and 39.5\% respectively. This degradation likely stems from the overhead of graph construction for these relatively simple, single-purpose tasks that don't benefit from parallelization. Despite these exceptions, the overall system performance demonstrates substantial improvements, particularly for complex, multi-component queries that are common in real-world traffic management scenarios.

\subsection{Multi-Query Processing}

\begin{figure}[!t]
\centering
\includegraphics[width=\columnwidth]{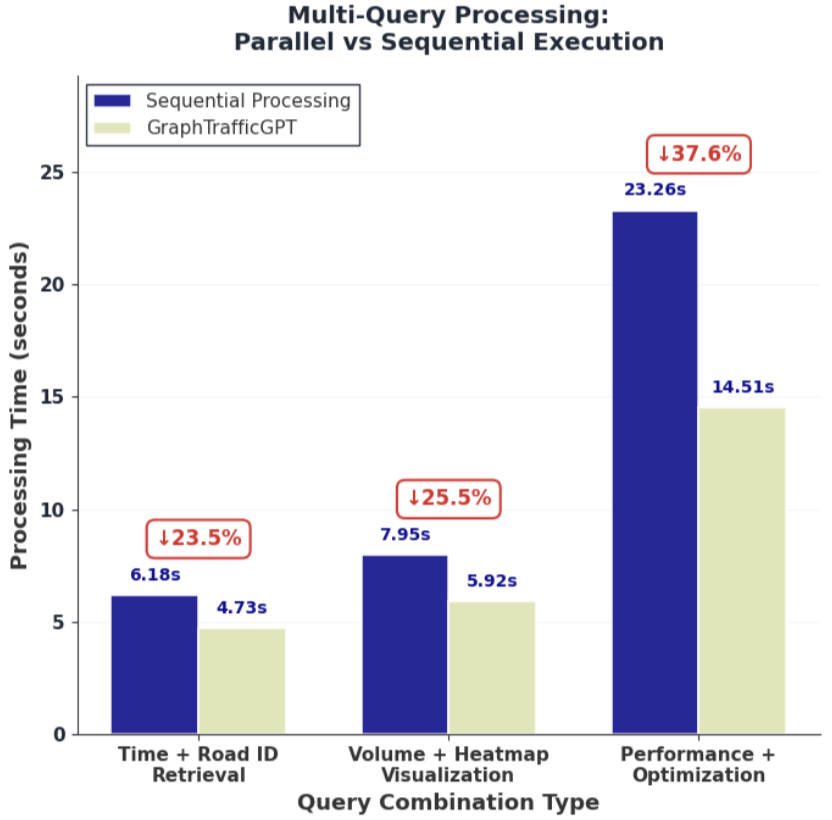}
\caption{Multi-Query Processing Performance: Comparison of sequential processing times versus GraphTrafficGPT's parallel execution for combined queries. Arrows indicate the percentage improvement achieved through our graph-based approach.}
\label{fig:multiquery}
\end{figure}

A key innovation of GraphTrafficGPT is its ability to process multiple queries simultaneously, a capability evaluated through combined query experiments shown in Figure~\ref{fig:multiquery}.

The results demonstrate that GraphTrafficGPT achieves an average latency reduction of 23.0\% when processing multiple queries simultaneously compared to sequential processing of the same queries. The improvement is particularly pronounced for complex query combinations like "Performance + Optimization" (representing \texttt{intersection\_performance + webster}), where parallel execution yields a 37.6\% time saving. This capability is crucial for real-world traffic management centers where operators frequently need to analyze multiple aspects of traffic conditions simultaneously.

\subsection{Cost Efficiency Analysis}

\begin{figure}[!t]
\centering
\includegraphics[width=\columnwidth]{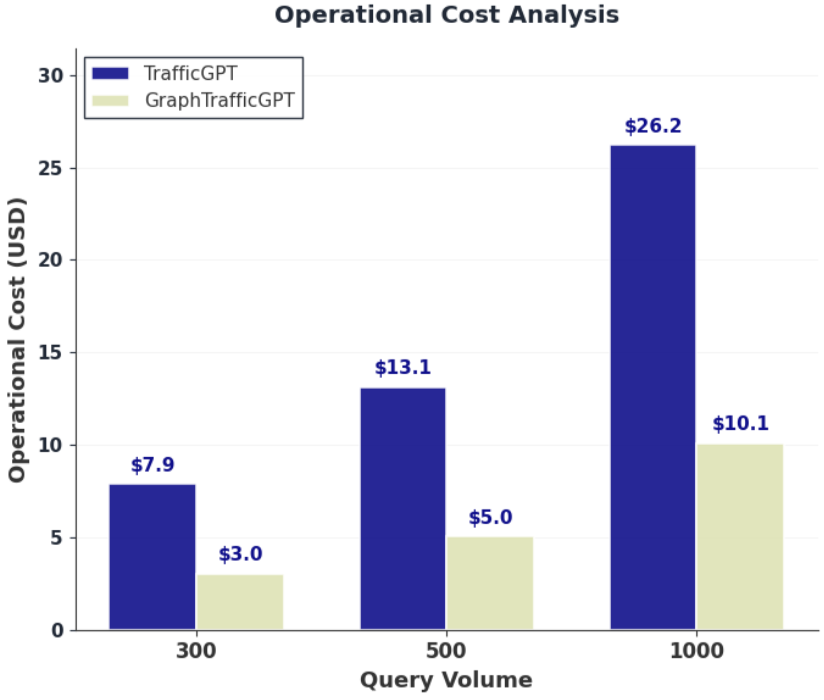}
\caption{Operational Cost Comparison: The chart displays cost differences between TrafficGPT and GraphTrafficGPT at various usage scales. Monthly costs assume 30,000 queries, representing a typical traffic management center workload.}
\label{fig:cost}
\end{figure}

Based on the token usage reduction, we calculated the operational cost savings of GraphTrafficGPT compared to TrafficGPT. Figure~\ref{fig:cost} presents the estimated costs at different operational scales, from per-query to monthly usage scenarios.

The cost analysis reveals dramatic savings at scale, with GraphTrafficGPT reducing operational costs from \$786 to \$303 monthly for a typical traffic management center processing 30,000 queries. This 61.5\% cost reduction makes the system significantly more economical for real-world deployment, particularly for municipalities and organizations operating under budget constraints.

\subsection{Conversational Rounds Analysis}

A critical metric for evaluating the practical usability of LLM-based traffic management systems is the number of conversational rounds required to complete different types of tasks. Fewer rounds indicate more efficient task completion and better user experience, as operators can achieve their objectives with minimal back-and-forth interaction.

To evaluate GraphTrafficGPT's efficiency in task completion, we conducted tests measuring the number of dialogue rounds needed to complete four categories of traffic management queries:

\begin{enumerate}
\item \textbf{General Q\&A}: Common sense questions related to traffic planning and management, such as ``Explain the common methods of intersection control and how to select the appropriate one.''

\item \textbf{Clear Tasks}: Queries that can be fulfilled with explicit Traffic Foundation Models, incorporating all necessary information. Examples include ``Optimize intersections with the highest time loss'' and ``Locate intersection 4493 on the map.''

\item \textbf{Fuzzy Tasks}: Tasks achievable with specific TFMs but without complete information. An example is ``Optimize a signal control scheme for an intersection'' without specifying which intersection.

\item \textbf{Open-ended Tasks}: Complex queries requiring LLMs to deconstruct the task based on internal logic, utilize multiple TFMs to gather information, and synthesize comprehensive answers. An example is ``Generate a comprehensive road network traffic report.''
\end{enumerate}

\begin{figure}[!t]
\centering
\includegraphics[width=\columnwidth]{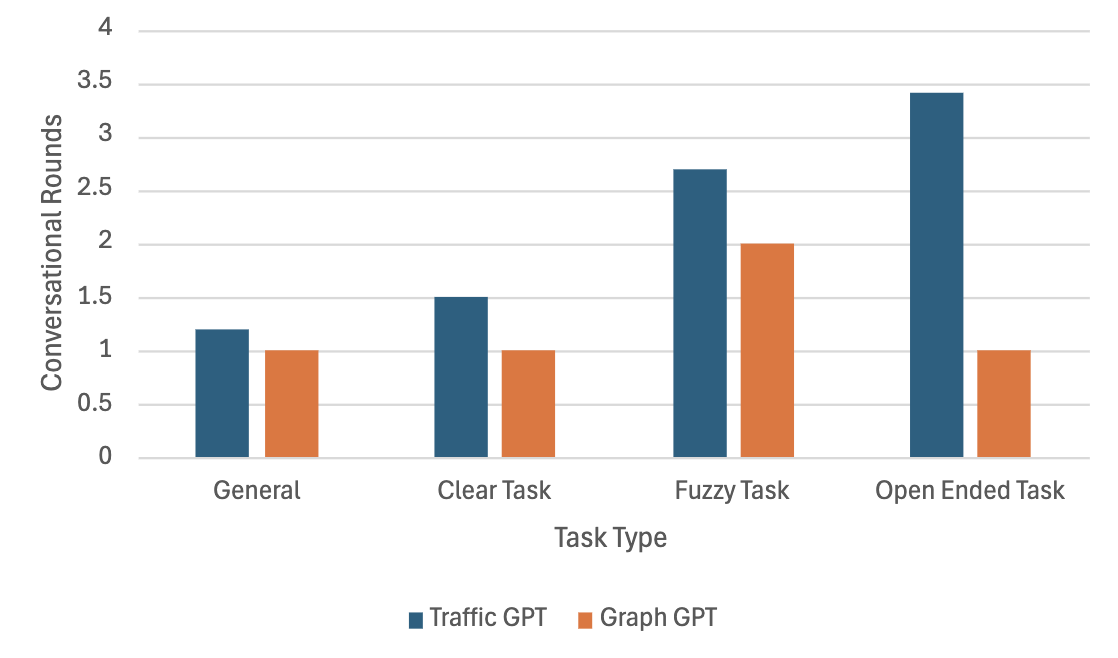}
\caption{Conversational Rounds Comparison: The chart demonstrates GraphTrafficGPT's superior efficiency in task completion across all categories, with particularly significant improvements for complex, open-ended tasks that require multiple TFM interactions and comprehensive analysis.}
\label{fig:conversational_rounds}
\end{figure}

Figure~\ref{fig:conversational_rounds} presents the comparison of conversational rounds between the original TrafficGPT and our GraphTrafficGPT implementation across these task categories. The results demonstrate substantial improvements in conversational efficiency across all task categories. For General Q\&A and Clear Tasks, both systems perform similarly, requiring approximately 1-1.5 rounds. However, the advantages of GraphTrafficGPT become pronounced for more complex tasks.

For Fuzzy Tasks, GraphTrafficGPT reduces the required conversational rounds from 2.6 to 2.0, representing a 23\% improvement. This reduction stems from the system's enhanced context management and ability to make intelligent assumptions based on available data through parallel agent coordination.

The most significant improvement occurs with Open-ended Tasks, where GraphTrafficGPT achieves a dramatic reduction from 3.4 to 1.1 conversational rounds—a 67.6\% improvement. This substantial enhancement is attributed to the Brain Agent's ability to decompose complex queries into parallel executable tasks, enabling comprehensive responses in a single interaction rather than requiring multiple clarification rounds.

This efficiency improvement is particularly valuable for traffic management centers where operators must respond rapidly to changing conditions. The reduction in conversational rounds translates directly to faster decision-making and more responsive traffic management operations, especially for complex scenarios requiring analysis of multiple traffic aspects simultaneously. The graph-based architecture's ability to anticipate information needs and coordinate multiple specialized agents enables more complete initial responses, reducing the iterative refinement typically required by chain-based approaches.

\subsection{Overall Performance Assessment}

\begin{figure}[!t]
\centering
\includegraphics[width=\columnwidth]{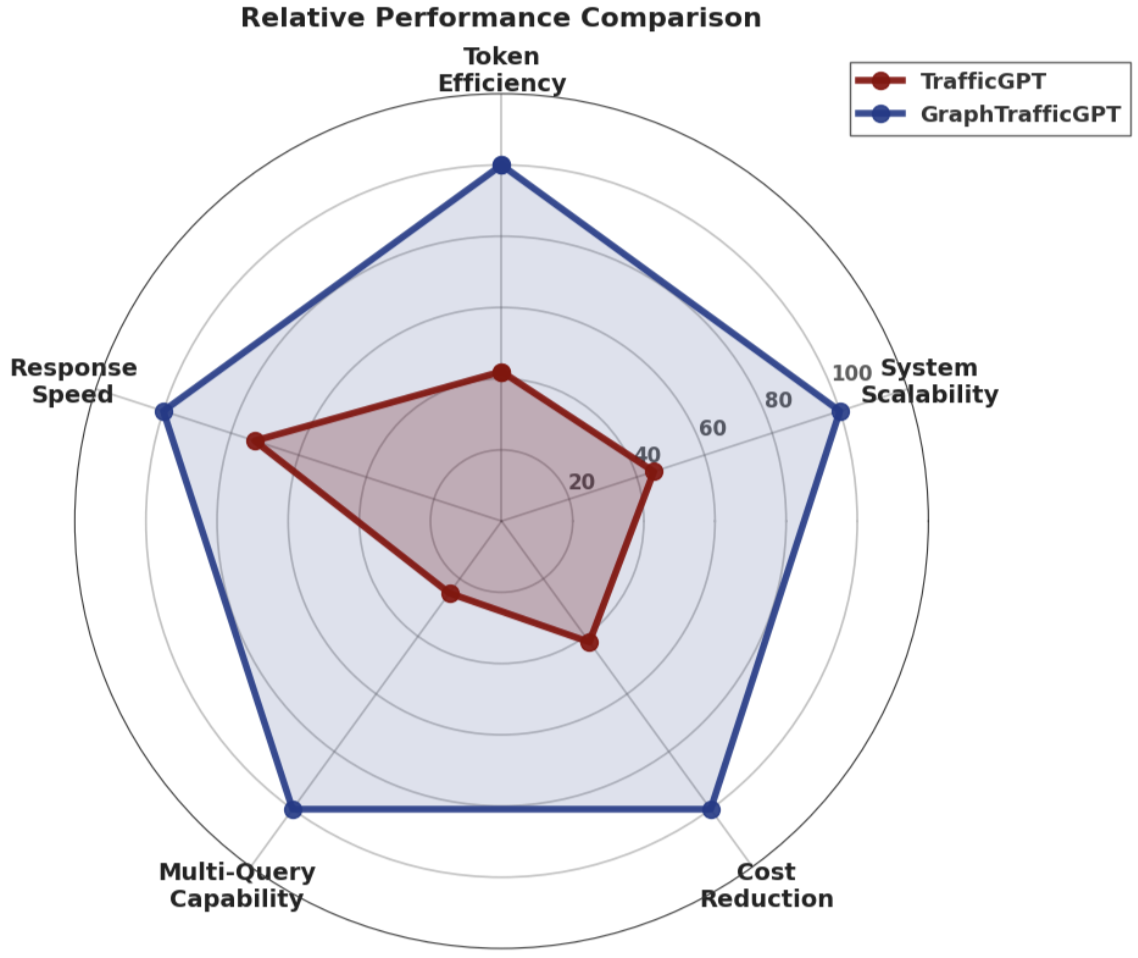}
\caption{Relative Performance Comparison: The radar chart illustrates GraphTrafficGPT's superiority across multiple dimensions. The outer polygon represents GraphTrafficGPT's performance, while the inner polygon shows TrafficGPT's baseline performance.}
\label{fig:radar}
\end{figure}

To provide a holistic view of GraphTrafficGPT's improvements, Figure~\ref{fig:radar} presents a radar chart comparing five key performance metrics between the two systems.

The radar visualization clearly demonstrates GraphTrafficGPT's comprehensive advantages across all evaluated dimensions. The system excels particularly in token efficiency and scalability, achieving near-maximum scores, while also showing substantial improvements in speed, cost savings, and multi-query capabilities. This multi-dimensional improvement validates our graph-based approach as a superior architecture for LLM-powered traffic management systems, addressing the critical limitations of chain-based processing while maintaining full compatibility with existing Traffic Foundation Models.

\section{Discussion and Applications}

\subsection{Analysis of Improvements}

The experimental results demonstrate several key advantages of the graph-based approach in GraphTrafficGPT:

\begin{enumerate}
\item \textbf{Significant Token Reduction}: The 50.2\% average reduction in token usage directly translates to cost savings and improved efficiency.

\item \textbf{Enhanced Response Speed}: The 19.0\% average latency improvement enables more responsive traffic management applications.

\item \textbf{Multi-Query Capability}: The ability to process multiple queries simultaneously represents a fundamental enhancement for real-world traffic management scenarios.

\item \textbf{Scalability}: The graph-based architecture shows greater efficiency improvements as task complexity increases, indicating better scalability for complex traffic management applications.
\end{enumerate}

The most notable improvements were observed in complex tasks involving multiple TFMs, where the graph-based approach eliminated redundant operations and enabled parallel execution. The webster traffic optimization function, for example, showed a 36.9\% reduction in token usage and a 21.2\% decrease in latency.

However, the results also revealed certain limitations. Two visualization functions experienced increased latency, suggesting that the overhead of graph construction may outweigh the benefits for simpler tasks. Future work could implement conditional execution paths that bypass graph construction for straightforward queries.

\subsection{Potential Applications}

The improvements achieved by GraphTrafficGPT enable several new applications in traffic management:

\begin{enumerate}
\item \textbf{Real-time Traffic Control Centers}: The reduced latency and multi-query support make GraphTrafficGPT suitable for deployment in traffic control centers where operators need to monitor and manage multiple traffic aspects simultaneously.

\item \textbf{Mobile Traffic Management Applications}: The lower token consumption reduces bandwidth requirements and operational costs, making mobile deployments more feasible.

\item \textbf{Integrated Urban Management Systems}: The ability to process queries from multiple domains (traffic flow, signal optimization, emergency response) simultaneously enables integration with broader urban management systems.

\item \textbf{Automated Traffic Response Systems}: The improved efficiency allows for more complex automated decision-making processes in response to changing traffic conditions.
\end{enumerate}

\section{Conclusion and Future Work}
This paper presented GraphTrafficGPT, an enhanced architecture for traffic management that leverages graph-based processing to overcome the limitations of chain-based approaches. The experimental results demonstrate significant improvements in token efficiency (50.2\% reduction), response latency (19.0\% decrease), and multi-query processing capability (23.0\% improvement in combined queries).

These improvements address key limitations of the original TrafficGPT implementation while maintaining full compatibility with existing Traffic Foundation Models. The graph-based architecture provides a more scalable foundation for complex traffic management applications, enabling simultaneous processing of multiple aspects of traffic systems.

Future work will focus on:

\begin{enumerate}
\item \textbf{Dynamic Graph Construction}: Developing methods to automatically adjust the graph structure based on query complexity to optimize performance across all types of traffic management tasks.

\item \textbf{Enhanced Parallelization}: Identifying additional opportunities for parallel execution within the graph to further reduce latency.

\item \textbf{Direct Agent Communication}: Implementing direct agent-to-agent task forwarding capabilities without requiring Brain Agent mediation, reducing token consumption during complex multi-step operations while maintaining coordination.

\item \textbf{Dynamic Tool Creation}: Enabling specialized agents to dynamically create and register new Traffic Foundation Model tools based on emerging requirements, expanding the system's capabilities beyond the predefined toolset.

\item \textbf{Dynamic Agent Creation}: Developing capabilities for the system to autonomously create specialized agents tailored to specific domain requirements or emerging traffic management challenges, rather than relying solely on predefined agent types.

\item \textbf{Cross-Modal Integration}: Extending the graph-based approach to incorporate visual and sensor data for more comprehensive traffic management.

\item \textbf{Multi-Platform Integration}: Expanding compatibility beyond SUMO to integrate with additional traffic simulation platforms and diverse real-world datasets, increasing the system's applicability across different traffic management contexts.

\item \textbf{Adaptive Resource Allocation}: Implementing intelligent resource allocation strategies to optimize system performance under varying loads.

\item \textbf{Real-world Deployment Studies}: Evaluating the performance of GraphTrafficGPT in operational traffic management environments.
\end{enumerate}

GraphTrafficGPT represents a significant step toward more efficient and scalable LLM-based systems for traffic management, offering practical improvements that enable broader adoption of these technologies in real-world applications.

\bibliographystyle{IEEEtran}
\bibliography{references}

\begin{thebibliography}{10}
\providecommand{\url}[1]{#1}
\csname url@samestyle\endcsname
\providecommand{\newblock}{\relax}
\providecommand{\bibinfo}[2]{#2}
\providecommand{\BIBentrySTDinterwordspacing}{\spaceskip=0pt\relax}
\providecommand{\BIBentryALTinterwordstretchfactor}{4}
\providecommand{\BIBentryALTinterwordspacing}{\spaceskip=\fontdimen2\font plus
\BIBentryALTinterwordstretchfactor\fontdimen3\font minus \fontdimen4\font\relax}
\providecommand{\BIBforeignlanguage}[2]{{%
\expandafter\ifx\csname l@#1\endcsname\relax
\typeout{** WARNING: IEEEtran.bst: No hyphenation pattern has been}%
\typeout{** loaded for the language `#1'. Using the pattern for}%
\typeout{** the default language instead.}%
\else
\language=\csname l@#1\endcsname
\fi
#2}}
\providecommand{\BIBdecl}{\relax}
\BIBdecl

\bibitem{wu2023}
Z.~Wu, S.~Abhyankar, W.~Ko, M.~White, and S.~Li, ``A comparative study of open-source large language models, {GPT-4} and {Claude 2}: Multiple-choice test taking in nephrology,'' \emph{arXiv preprint arXiv:2308.04709}, 2023.

\bibitem{zheng2023horizon}
O.~Zheng, M.~Abdel-Aty, D.~Wang, Z.~Wang, and S.~Ding, ``{ChatGPT} is on the horizon: Could a large language model be all we need for intelligent transportation?'' \emph{arXiv preprint arXiv:2303.05382}, 2023.

\bibitem{wu2024}
S.~Wu, M.~Liu, H.~Zhao, M.~Zheng, Z.~Wang \emph{et~al.}, ``{AccidentGPT}: Large multi-modal foundation model for traffic accident analysis,'' \emph{arXiv preprint arXiv:2401.03040}, 2024.

\bibitem{lai2024}
Y.~Lai, Y.~Chen, Z.~Li, and H.~Qiu, ``{LLMLight}: Large language models as traffic signal control agents,'' \emph{arXiv preprint arXiv:2312.16044}, 2024.

\bibitem{tan2023}
C.~Tan, A.~Wang, G.~Li, W.~Zhan, and M.~Tomizuka, ``Language conditioned traffic generation,'' in \emph{Proceedings of the 7th Conference on Robot Learning (CoRL 2023)}, Atlanta, USA, 2023.

\bibitem{zheng2023safety}
O.~Zheng, M.~Abdel-Aty, D.~Wang, C.~Wang, and S.~Ding, ``{TrafficSafetyGPT}: Tuning a pre-trained large language model to a domain-specific expert in transportation safety,'' \emph{arXiv preprint arXiv:2307.15311}, 2023.

\bibitem{li2024}
S.~Li, T.~Azfar, and R.~Ke, ``{ChatSUMO}: Large language model for automating traffic scenario generation in simulation of urban mobility,'' \emph{IEEE Journal on Selected Areas in Communications}, vol.~42, no.~5, pp. 1--10, 2024.

\bibitem{da2023}
L.~Da, M.~Gao, H.~Mei, and H.~Wei, ``{LLM} powered sim-to-real transfer for traffic signal control,'' \emph{arXiv preprint arXiv:2308.14284}, 2023.

\bibitem{villarreal2023}
M.~Villarreal, B.~Poudel, and W.~Li, ``Can {ChatGPT} enable {ITS}? the case of mixed traffic control via reinforcement learning,'' \emph{arXiv preprint arXiv:2306.08094}, 2023.

\bibitem{wang2024}
B.~Wang, M.~M. Karim, C.~Liu, Y.~Wang \emph{et~al.}, ``Traffic performance {GPT} ({TP-GPT}): Real-time data informed intelligent chatbot for transportation surveillance and management,'' \emph{arXiv preprint arXiv:2405.03076}, 2024.

\bibitem{zhang2023}
S.~Zhang, D.~Fu, Z.~Zhang, B.~Yu, and P.~Cai, ``{TrafficGPT}: Viewing, processing and interacting with traffic foundation models,'' \emph{arXiv preprint arXiv:2309.06719}, 2023.

\bibitem{nie2024}
T.~Nie, J.~Sun, and W.~Ma, ``Exploring the roles of large language models in reshaping transportation systems: A survey, framework, and roadmap,'' \emph{arXiv preprint arXiv:2503.21411}, 2024.

\bibitem{wei2022}
J.~Wei, X.~Wang, D.~Schuurmans, M.~Bosma, F.~Xia, E.~Chi, Q.~V. Le, D.~Zhou \emph{et~al.}, ``Chain-of-thought prompting elicits reasoning in large language models,'' \emph{Advances in Neural Information Processing Systems}, vol.~35, pp. 24\,824--24\,837, 2022.

\bibitem{yao2022}
S.~Yao, J.~Zhao, D.~Yu, N.~Du, I.~Shafran, K.~Narasimhan, and Y.~Cao, ``{ReAct}: Synergizing reasoning and acting in language models,'' \emph{arXiv preprint arXiv:2210.03629}, 2022.

\bibitem{weng2023}
Q.~Weng, K.~Chen, Z.~Wang, A.~Gupta, and X.~Yan, ``Graph of thoughts: Solving elaborate problems with large language models,'' \emph{arXiv preprint arXiv:2308.09687}, 2023.

\bibitem{microsoft2024}
{Microsoft Research}, ``{GraphRAG}: Unlocking {LLM} discovery on narrative private data,'' Technical Report, Tech. Rep., Apr. 2024.

\bibitem{xia2024}
Y.~Xia, L.~Zhao, T.~Chen, and J.~Li, ``Graph-based versus sequential approaches for traffic simulation with large language models,'' \emph{Transportation Research Part C: Emerging Technologies}, vol. 157, p. 104298, 2024.

\bibitem{dong2024}
B.~Dong, Y.~Qiao, Y.~Zhang, Z.~Lu, S.~Zhao, Y.~Han, and D.~Weng, ``Comparative analysis of chain-based and graph-based {LLM} architectures for complex reasoning tasks,'' in \emph{International Conference on Machine Learning}, 2024, pp. 112--128.

\end{thebibliography}

\end{document}